\documentclass[11pt,letterpaper]{article}
\usepackage{emnlp2016}
\usepackage{times}

\usepackage{latexsym}

\emnlpfinalcopy

\def\emnlppaperid{***}

\usepackage{todonotes}
\usepackage[hidelinks]{hyperref}
\usepackage{url}
\usepackage{paralist}
\usepackage{amsmath,amsthm}
\usepackage{float}
\usepackage{array}
\usepackage{algorithm}
\usepackage{algorithmic}
\usepackage{graphicx}
\usepackage{multirow}

\DeclareMathOperator{\T}{\mathcal{T}}
\DeclareMathOperator{\Expr}{\textsc{Expr}}
\DeclareMathOperator{\Op}{\odot}
\DeclareMathOperator{\Order}{\textsc{Order}}

\DeclareMathOperator{\Relevance}{\mathrm{Relevance}}
\DeclareMathOperator{\Location}{\mathrm{Location}}
\DeclareMathOperator{\SpanStart}{\mathrm{Span-Start}}
\DeclareMathOperator{\SpanEnd}{\mathrm{Span-End}}
\DeclareMathOperator{\MidSpan}{\mathrm{MidSpan}}
\DeclareMathOperator{\LeftSpan}{\mathrm{LeftSpan}}
\DeclareMathOperator{\RightSpan}{\mathrm{RightSpan}}
\DeclareMathOperator{\LeftToken}{\mathrm{LeftToken}}

\newcolumntype{L}[1]{>{\raggedright\let\newline\\\arraybackslash\hspace{0pt}}m{#1}}
\newcolumntype{C}[1]{>{\centering\let\newline\\\arraybackslash\hspace{0pt}}m{#1}}
\newcolumntype{R}[1]{>{\raggedleft\let\newline\\\arraybackslash\hspace{0pt}}m{#1}}

\title{\textsc{Equation Parsing} : Mapping Sentences to Grounded Equations}

\author{Subhro Roy \quad\quad Shyam Upadhyay \quad\quad Dan Roth \\
 University of Illinois, Urbana Champaign \\
 {\tt \{sroy9, upadhya3, danr\}@illinois.edu} \\}

\date{}

\begin{document}
\maketitle
\begin{abstract}
  Identifying mathematical relations expressed in text is essential to
  understanding a broad range of natural language text from election
  reports, to financial news, to sport commentaries to mathematical
  word problems. This paper focuses on identifying and understanding
  mathematical relations described within a single sentence. We
  introduce the problem of Equation Parsing -- given a sentence,
  identify noun phrases which represent variables, and generate the
  mathematical equation expressing the relation described in the
  sentence. We introduce the notion of projective equation parsing and
  provide an efficient algorithm to parse text to projective
  equations. Our system makes use of a high precision lexicon of
  mathematical expressions and a pipeline of structured predictors,
  and generates correct equations in $70\%$ of the cases. In $60\%$ of
  the time, it also identifies the correct noun phrase $\rightarrow$
  variables mapping, significantly outperforming baselines. We also
  release a new annotated dataset for task evaluation.

\end{abstract}

\section{Introduction}
\label{sec:intro}

  Understanding text often involves reasoning with respect to
  quantities mentioned in it. Understanding the news article statement
  in Example 1 requires identifying relevant entities and the
  mathematical relations expressed among them in text, and determining
  how to compose them. Similarly, solving a math word problem with a
  sentence like Example 2, requires realizing that it deals with a
  {\em single} number, knowing the meaning of {\em``difference''} and
  composing the right equation -- {\em``25''} needs to be subtracted
  from a number only after it is multiplied by $3$.

  \begin{table}
  \centering
  \begin{tabular}{|p{7cm}|} \hline

    \textbf{Example 1} {\em Emanuel's campaign contributions total
    three times those of his opponents put together.} \\\hline

    \textbf{Example 2} {\em Twice a number equals 25 less than triple
         the same number.} \\\hline

    \textbf{Example 3} {\em Flying with the wind , a bird was able to
    make 150 kilometers per hour.} \\\hline

    \textbf{Example 4} {\em The sum of two numbers is 80.}\\\hline

    \textbf{Example 5} {\em There are 54 5-dollar and 10-dollar
    notes.}\\\hline

  \end{tabular}      
  \end{table}

  As a first step towards understanding such relations, we introduce
  the Equation Parsing task - given a sentence expressing a
  mathematical relation, the goal is to generate an equation
  representing the relation, and to map the variables in the equation
  to their corresponding noun phrases. To keep the problem tractable,
  in this paper we restrict the final output equation form to have at
  most two (possibly coreferent) variables, and assume that each
  quantity mentioned in the sentence can be used at most once in the
  final equation.\footnote{We empirically found that around 97\% of
  sentences describing a relation have this property.} In example 1,
  the gold output of an equation parse should be $V_1 = 3 \times V_2$,
  with $V_1=\text{``Emanuel's campaign contributions''}$ and
  $V_2=\text{``those of his opponents put together''}$.

  The task can be seen as a form of semantic
  parsing \cite{GoldwasserRo11,KCAZ13} where instead of mapping a
  sentence to a logical form, we want to map it to an
  equation. However, there are some key differences that make this
  problem very challenging in ways that differ from the ``standard''
  semantic parsing.  In Equation Parsing, not all the components of
  the sentence are mapped to the final equation. There is a need to
  identify noun phrases that correspond to variables in the relations
  and determine that some are irrelevant and can be dropped. Moreover,
  in difference from semantic parsing into logical forms, in Equation
  Parsing multiple phrases in the text could correspond to the same
  variable, and identical phrases in the text could correspond to
  multiple variables.

  We call the problem of mapping noun phrases to variables the problem
  of {\em grounding variables}. Grounding is challenging for various
  reasons, key among them are that: (i) The text often does not
  mention {\em``variables''} explicitly, e.g., the sentence in example
  3 describes a mathematical relation between the speed of bird and
  the speed of wind, without mentioning ``speed'' explicitly. (ii)
  Sometimes, multiple noun phrases could refer to the same
  variable. For instance, in example 2, both {\em``a number''} and
  {\em``the same number''} refer to the same variable. On the other
  hand, the same noun phrase might refer to multiple variables, as in
  example 4, where the noun phrase ``two numbers'' refer to two variables.

  In addition, the task involves deciding which of the quantities
  identified in the sentence are relevant to the final equation
  generation. In example 5, both ``5'' and ``10'' are not relevant for
  the final equation ``$V_1+V_2=54$''. Finally, the equation needs to
  be constructed from a list of relevant quantities and grounded
  variables. Overall, the output space becomes exponential in the
  number of quantities mentioned in the sentence.

  Determining the final equation that corresponds to the text is an
  inference step over a very large space. To address this, we define
  the concept of ``projectivity'' - a condition where the final
  equation can be generated by combining adjacent numbers or
  variables, and show that most sentences expressing mathematical
  relations exhibit the projectivity property. Finally, we restrict
  our inference procedure to only search over equations which have
  this property.

  Our approach builds on a pipeline of structured predictors that
  identify irrelevant quantities, recognize coreferent variables, and,
  finally, generate equations. We also leverage a high precision
  lexicon of mathematical expressions and develop a greedy lexicon
  matching strategy to guide inference. We discuss and exemplify the
  advantages of this approach and, in particular, explain where the
  ``standard'' NLP pipeline fails to support equation parsing, and
  necessitates the new approach proposed here. Another contribution of
  this work is the development of a new annotated data set for the
  task of equation parsing. We evaluate our method on this dataset and
  show that our method predicts the correct equation in $70\%$ of the
  cases and that in $60\%$ of the time we also ground all variables
  correctly.

  The next section presents a discussion of related work. Next we
  formally describe the task of equation parsing. The following
  sections describe our equation representation and the concept of
  projectivity, followed by the description of our algorithm to
  generate the equations and variable groundings from text. We
  conclude with experimental results.

\section{Related Work}
  
  The work most related to this paper is \cite{MMMRS16}, which focuses
  on extracting relation triples where one of the arguments is a
  number. In contrast, our work deals with multiple variables and
  complex equations involving them. There has been a lot of recent
  work in automatic math word problem
  solving \cite{KushmanZeBa14,RoyViRo15,HosseiniHaEt14,RoyRo15}.
  These solvers cannot handle sentences individually. They require the
  input to be a complete math word problem, and even then, they only
  focus on retrieving a set of answer values without mentioning what
  each answer value corresponds to.
  Our work is also conceptually related
  to work on semantic parsing -- mapping natural language text to a
  formal meaning
  representation \cite{WongMo07,CGCR10,CaiYa13,KCAZ13,GoldwasserRo11}. However,
  as mentioned earlier, there are some significant differences in the
  task definition that necessitate the development of a new approach.

\section{The Equation Parsing Task}

  Equation parsing takes as input a sentence $x$ describing a single
  mathematical equation, comprising one or two variables and other
  quantities mentioned in $x$. Let $N$ be the set of noun phrases in
  the sentence $x$. The output of the task is the mathematical
  equation described in $x$, along with a mapping of each variable in
  the equation to its corresponding noun phrase in $N$. We refer to
  this mapping as the ``grounding'' of the variable; the noun phrase
  represents what the variable stands for in the equation.
  Table \ref{tab:inputoutput} gives an example of an input and output
  for the equation parsing of the text in example 2. Since an equation
  can be written in various forms, we use the form which most agrees
  with text, as our target output. So, for example 1, we will choose
  $V_1=3 \times V_2$ and not $V_2=V_1 \div 3$. In cases where several
  equation forms seem to be equally likely to be the target equation,
  we randomly choose one of them, and keep this choice consistent
  across the dataset.

  \setlength{\tabcolsep}{6pt}
  \begin{table}[H]
         \centering
         \small
         \begin{tabular}{|lp{5cm}|} \hline
         \multicolumn{2}{|l|}{The Equation Parsing Task}\\ \hline
         Input & {\em Twice a number equals 25 less than triple 
         the same number.} \\ \hline
         Output & $2 \times V_1=(3 \times V_1) - 25$ {\em(Equation)}\\ 
         & $V_1= \text{``a number''}$ {\em(Grounding)}\\   
        \hline                                                                
        \end{tabular}
        \caption{Input and output for Equation Parsing}
        \label{tab:inputoutput}
  \end{table}

\subsection{Equation Parse Representation}
  \begin{figure*}
    \begin{center}
    \frame{\includegraphics[width=0.8\textwidth]{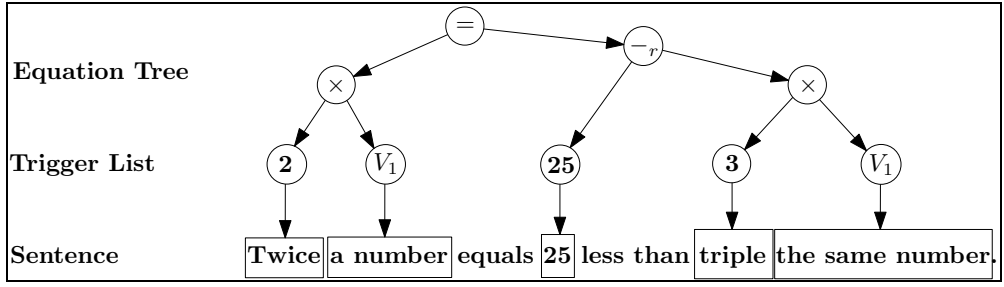}}
    \end{center}
    \caption{A sentence with its trigger list and 
      equation tree. $-_r$ indicates subtraction with order $rl$.}
    \label{fig:eqparse}
  \end{figure*}  
  In this section, we introduce an equation parse for a sentence. An
  equation parse of a sentence $x$ is a pair $(T, E)$, where $T$
  represents a set of {\em triggers} extracted from $x$, and $E$
  represents an {\em equation tree} formed with the set $T$ as
  leaves. We now describe these terms in detail.
  
  \noindent \textbf{Trigger} Given a sentence $x$ mentioning a
  mathematical relation, a trigger can either be a {\em quantity
    trigger} expressed in $x$, or {\em variable trigger} which is a
  noun phrase in $x$ corresponding to a variable. A {\em quantity
    trigger} is a tuple $(q, s)$, where $q$ is the numeric value of
  the quantity mentioned in text, and $s$ is the span of text from the
  sentence $x$ which refers to the quantity. A {\em variable trigger}
  is a tuple $(l, s)$, where $l$ represents the label of the variable,
  and $s$ represents the noun phrase representing the variable. For
  example, for the sentence in Fig \ref{fig:eqparse}, the spans
  ``Twice'', ``25'', and ``triple'' generate quantity triggers,
  whereas ``a number'' and ``the same number'' generate variable
  triggers, with label $V_1$.

  \noindent \textbf{Trigger List} The trigger list $T$ for a sentence
  $x$ contains one trigger for each variable mention and each numeric
  value used in the final equation expressed by the sentence $x$. The
  trigger list might consist of multiple triggers having the same
  label, or extracted from the same span of text.  In the example
  sentence in Fig \ref{fig:eqparse}, the trigger list comprises two
  triggers having the same label $V_1$. The final trigger list for the
  example in Fig \ref{fig:eqparse} is \{($2$, ``2''), ($V_1$, ``a
  number''), ($25$, ``25''), ($3$, ``triple''), ($V_1$, ``the same
  number'')\}. Note that there can be multiple valid trigger lists. In
  our example, we could have chosen both variable triggers to point to
  the same mention ``a number''. Quantity triggers in the trigger list
  form the \textbf{\em quantity trigger list}, and the variable
  triggers in trigger list form the \textbf{\em variable trigger
    list}.
  \begin{table}
    \footnotesize
    \centering
    \begin{tabular}{|p{2.2cm}|p{4.7cm}|}\hline
      Notation & Definition \\\hline\hline
      Quantity Trigger & Mention of a quantity in text\\\hline
      Variable Trigger & Noun phrase coupled with variable label\\\hline
      Trigger & Quantity or variable trigger\\\hline
      Quantity Trigger List & List of quantity triggers, one for each number mention in equation\\\hline
      Variable Trigger List & List of variable triggers, one for each variable mention in equation\\\hline
      Trigger List & Union of quantity and variable trigger list\\\hline
      Equation Tree & Binary tree representation of equation\\\hline
      $lc(n)$, $rc(n)$ & Left and right child of node $n$\\\hline
      $\Expr(n)$ & Expression represented by node $n$\\\hline
      $\Op(n)$ & Operation at node $n$\\\hline
      $\Order(n)$ & Order of operation at node $n$\\\hline
      $\Location(n)$ & Character offset of trigger representing leaf node $n$\\\hline
      $\SpanStart(n)$, $\SpanEnd(n)$ & Start and end character offsets of span covered by
      node $n$\\\hline
    \end{tabular}
    \label{tab:notation}
    \caption{Summary of notations used in the paper}
  \end{table}  

  \noindent \textbf{Equation Tree} An equation tree of a sentence $x$
  is a binary tree whose leaves constitute the trigger list of $x$,
  and internal nodes (except the root) are labeled with one of the
  following operations -- {\em addition, subtraction, multiplication,
    division}. In addition, for nodes which are labeled with
  subtraction or division, we maintain a separate variable to
  determine order of its children. The root of the tree is always
  labeled with the operation {\em equal}.

  An equation tree is a natural representation for an equation. Each
  node $n$ in an equation tree represents an expression $\Expr(n)$,
  and the label of the parent node determines how the expressions of
  its children are to be composed to construct its own expression. Let
  us denote the label for a non-leaf node $n$ to be $\Op(n)$, where
  $\Op(n) \in \{+, -, \times, \div, =\}$ and the order of a node $n$'s
  children by $\Order(n)$ (defined only for subtraction and division
  nodes), which takes values $lr$ (Left-Right) or $rl$ (Right-Left).
  For a leaf node $n$, the expression $\Expr(n)$ represents the
  variable label, if $n$ is a variable trigger, and the numeric value
  of the quantity, if it is a quantity trigger. Finally, we use
  $lc(n)$ and $rc(n)$ to represent the left and right child of node
  $n$, respectively. The equation represented by the tree can be
  generated as follows. For all non-leaf nodes $n$, we have
  \vspace{-0.1in}
  \begin{multline}\label{eq:parse}
    \Expr(n) = \\
    \begin{cases}
    \Expr(lc(n)) \Op(n) \Expr(rc(n)) \\
    \qquad \mbox{if } \Op(n) \in \{+,\times, =\} \\
    \Expr(lc(n)) \Op(n) \Expr(rc(n)) \\
    \qquad \mbox{if } \Op(n) \in \{-,\div\} \wedge \Order(n)=lr \\
    \Expr(rc(n)) \Op(n) \Expr(lc(n)) \\
    \qquad \mbox{if } \Op(n) \in \{-,\div\} \wedge \Order(n)=rl \\
    \end{cases}
  \end{multline}
  Given an equation tree $\T$ of a sentence, the equation represented
  by it is the expression generated by the root of $\T$ (following
  Equation \ref{eq:parse}). Referring to the equation tree in Fig
  \ref{fig:eqparse}, the node marked ``$-_r$'' represents
  $(3 \times V_1) - 25$, and the root represents the full equation
  $2 \times V_1 = (3 \times V_1)-25$.

\section{Projectivity}

  For each leaf $n$ of an equation tree $T$, we define a function
  $\Location(\cdot)$, to indicate the position of the corresponding
  trigger in text. We also define for each node $n$ of equation tree
  $T$, functions $\SpanStart(n)$ and $\SpanEnd(n)$ to denote the
  minimum span of text containing the leaves of the subtree rooted at
  $n$. We define them as follows:
  \vspace{-0.1in}
  \begin{multline}\label{eq:start}
    \SpanStart(n) = \\
    \begin{cases}
    \Location(n) \qquad \mbox{if $n$ is a leaf} \\
    \min(\SpanStart(lc(n)), \SpanStart(rc(n))) \\
    \qquad\qquad\qquad\qquad \mbox{otherwise}
    \end{cases}
  \end{multline}
  \vspace{-0.3in}
  \begin{multline}\label{eq:end}
    \SpanEnd(n) = \\
    \begin{cases}
    \Location(n) \qquad \mbox{if $n$ is a leaf} \\
    \max(\SpanEnd(lc(n)), \SpanEnd(rc(n))) \\
    \qquad\qquad\qquad\qquad \mbox{otherwise}
    \end{cases}
  \end{multline}
  An equation tree $T$ is called {\em projective} iff for every node
  $n$ of $T$, either $\SpanEnd(lc(n)) \leq \SpanStart(rc(n))$ or
  $\SpanEnd(rc(n)) \leq \SpanStart(lc(n))$.  In other words, the span
  of the left child and the right child cannot intersect in a
  projective equation tree\footnote{This is more general than the
  definition of projective trees used in dependency
  parsing \cite{MPRH05}.}.

  The key observation, as our corpus analysis indicates, is that for
  most sentences, {\em there exists} a trigger list, such that the
  equation tree representing the relation in the sentence is
  projective. However this might involve mapping two mentions of the
  same variable to different noun phrases. Figure \ref{eq:parse} shows
  an example of a projective equation tree, which requires different
  mentions of $V_1$ to be mapped to different noun phrases. If we had
  mapped both mentions of $V_1$ to same noun phrase ``a number'', the
  resulting equation tree would not have been projective. We collected
  $385$ sentences which represent an equation with one or two mentions
  of variables, and each number in the sentence used at most once in
  the equation.  We found that only one sentence among these could not
  generate a projective equation tree. (See Section \ref{sec:dataset}
  for details on dataset creation). Therefore, we develop an
  algorithmic approach for predicting projective equation trees, and
  show empirically that it compares favourably with ones which do not
  make the projective assumption.

\section{Predicting Equation Parse}

  Equation parsing of a sentence involves predicting three components
  -- Quantity Trigger List, Variable Trigger List and Equation Tree.
  We develop three structured prediction modules to predict each of
  the above components.

  All our prediction modules take a similar form: given input $x$ and
  output $y$, we learn a scoring function $f_w(x,y)$, which scores how
  likely is the output $y$ given input $x$. The scoring function
  $f_w(x,y)$ is linear, $f_w(y) = w^T\phi(x, y) $, where $\phi(x, y)$
  is a feature vector extracted from $x$ and $y$. The inference
  problem, that is, the prediction $y^*$ for an input $x$ is then:
  $y^* = \arg\max_{y\in \mathcal{Y}} f_w(y)$, where $\mathcal{Y}$ is
  the set of all allowed values of $y$.
   
  \subsection{Predicting Quantity Trigger List}
    
    Given input text and the quantities mentioned in it, the role of
    this step is to identify , for each quantity in the text, whether
    it should be part of the final equation. For instance, in example
    5 in Section \ref{sec:intro}, both ``5'' and ``10'' are not
    relevant for the final equation ``$V_1 + V_2 = 54$''. Similarly,
    in example 4, the number ``two'' is irrelevant for the equation
    ``$V_1+V_2=80$''.
    
    We define for each quantity $q$ in the sentence, a boolean value
    $\Relevance(q)$, which is set to $true$ if $q$ is relevant for the
    final equation, and to $false$ otherwise. For the structured
    classification, the input $x$ is the sentence along with a set of
    recognized quantities mentioned in it, and the output $y$ is the
    relevance values for all quantities in the sentence. We
    empirically found that predicting all relevance values jointly
    performs better than having a binary classifier predict each one
    separately. The feature function $\phi(x,y)$ used for the
    classification generates neighborhood features (from neighborhood
    of $q$) and quantity features (properties of the quantity
    mention). Details added to the appendix.

  \subsection{Predicting Variable Trigger List}

    The goal of this step is to predict the variable trigger list for
    the equation. Our structured classifier takes as input the
    sentence $x$, and the output $y$ is either one or two
    noun-phrases, representing variables in the final equation.  As we
    pointed out earlier, multiple groundings might be valid for any
    given variable, hence there can be multiple valid variable trigger
    lists. For every sentence $x$, we construct a set $Y$ of valid
    outputs. Each element in $Y$ corresponds to a valid variable
    trigger list. Finally, we aim to output only one of the elements
    of $Y$.

    We modified the standard structured prediction algorithm to
    consider ``superset supervision'' and take into account multiple
    gold structures for an input $x$. We assume access to $N$ training
    examples of the form : $(x_1, Y_1), (x_2, Y_2), \ldots, (x_N,
    Y_N)$, where each $Y_i$ is a set of valid outputs for the sentence
    $x_i$. Since we want to output only one variable trigger list, we
    want to score at least one $y$ from $Y_i$ higher than all other
    possible outputs, for each $x_i$.
    We use a modified
    latent structured SVM to learn the weight vector $w$. The 
    algorithm treats the best
    choice among all of $Y_i$ as a latent variable. At each
    iteration, for all $x_i$, the algorithm chooses the best choice
    $y_i^*$ from the set $Y_i$, according to the weight vector
    $w$. Then, $w$ is updated by learning on all $(x_i,y_i^*)$ by a
    standard structured SVM algorithm. The details of the algorithm
    are in Algorithm \ref{lssvm}.
      \begin{algorithm}
       \caption{Structural SVM with Superset Supervision}
       \label{lssvm}
       \begin{algorithmic}[1]
         \REQUIRE {Training data $T=\{(x_1, Y_1), (x_2, Y_2),
         \ldots, (x_N, Y_N)\}$} 
         \ENSURE Trained weight vector $w$
         \STATE $w \leftarrow w_0$
         \REPEAT 
           \STATE $T' \leftarrow \emptyset$
           \FORALL {$(x_i, Y_i) \in T$}
             \STATE $y_i^* \leftarrow \arg\max_{y \in Y_i} w^T\phi(x_i, y)$
             \STATE $T' \leftarrow T' \cup \{(x_i, y_i^*)\}$  
           \ENDFOR
           \STATE Update $w$ by running standard Structural SVM algorithm
           on $T'$
         \UNTIL{convergence}
         \RETURN $w$
       \end{algorithmic}
      \end{algorithm}
    The distinction from standard latent structural SVM is in line $5$
    of Algorithm \ref{lssvm}. In order to get the best choice $y_i^*$
    for input $x_i$, we search only inside $Y_i$, instead of all of
    $\mathcal{Y}$. A similar formulation can be found
    in \newcite{BjorkelundKu14}. The features $\phi(x,y)$ used for variable
    trigger prediction include variable features (properties of 
    noun phrase indicating variable) and neighborhood features (lexical
    features from neighborhood of variable mention).  Details added to 
    the appendix.

    If the output of the classifier is a pair of noun phrases, we use
    a rule based variable coreference detector, to determine whether
    both noun phrases should have the same variable label or
    not. The rules for variable coreference are as follows :
    \begin{enumerate}
      \item If both noun phrases are the same, and they do not have
        the token ``two'' or ``2'', they have the same label.
      \item If the noun phrases are different, and the noun phrase
        appearing later in the sentence contains tokens ``itself'',
        ``the same number'', they have the same label.
      \item In all other cases, they have different labels.
    \end{enumerate}  
    Finally, each noun phrase contributes one variable trigger to
    the variable trigger list.
 
  \subsection{Predicting Equation Tree} 

    It is natural to assume that the syntactic parse of the sentence
    could be very useful in addressing all the predictions we are
    making in the equation parsing tasks. However, it turns out that
    this is not the case -- large portions of the syntactic parse will
    not be part of the equation parse, hence we need the
    aforementioned modules to address this.  Nevertheless, in the next
    task of predicting the equation tree, we attempted to constraint
    the output space using guidance from the syntactic tree; we found,
    though, that even enforcing this weak level of output expectation
    is not productive. This was due to the poor performance of current
    syntactic parsers on the equation data (eg., in $32\%$ of
    sentences, the Stanford parser made a mistake which does not allow
    recovering the correct equation).

    The tree prediction module receives the trigger list predicted by
    the previous two modules, and the goal is to create an equation
    tree using the trigger list as the leaves of that tree. The input
    $x$ is the sentence and the trigger list, and the output $y$ is
    the equation tree representing the relation described in the
    sentence. We assume that the output will be a projective equation
    tree.  For features $\phi(x,y)$, we extract for each non-leaf node
    $n$ of the equation tree $y$, neighborhood features (from
    neighborhood of node spans of $n$'s children), connecting text
    features (from text between the spans of $n$'s children) and
    number features (properties of number in case of leaf nodes).
    Details are included in the appendix.

    The projectivity assumption implies that the final equation tree
    can be generated by combining only adjacent nodes, once the set of
    leaves is sorted based on $\SpanStart(\cdot)$ values. This allows
    us to use CKY algorithm for inference. A natural approach to
    further reduce the output space is to conform to the projective
    structure of the syntactic parse of the sentence. However, we
    found this to adversely affect performance, due to the poor
    performance of syntactic parser on equation data.

    \noindent \textbf{Lexicon} To bootstrap the equation parsing
    process, we developed a high precision lexicon to translate
    mathematical expressions to operations and orders, like ``sum of A
    and B'' translates to ``A+B'', ``A minus B'' translates to
    ``A-B'', etc. (where A and B denote placeholder numbers or
    expressions). At each step of CKY, while constructing a node $n$
    of the equation tree, we check for a lexicon text expression
    corresponding to node $n$. If found, we allow only the
    corresponding operation (and order) for node $n$, and do not
    explore other operations or orders. We show empirically that
    reducing the space using this greedy lexicon matching help improve
    performance. We found that using the lexicon rules as features
    instead of hard constraints do not help as much. Note that our
    lexicon comprises only generic math concepts, and around $50\%$ of
    the sentences in our dataset do not contain any pattern from the
    lexicon.
    



   Finally, given input sentence, we first predict the quantity
    trigger and the variable trigger lists. Given the complete trigger
    list, we predict the equation tree relating the components of the
    trigger list.
  
  \subsection{Alternatives} \label{sec:alternatives} A natural
  approach could be to jointly learn to predict all three components,
  to capture the dependencies among them.  
  To investigate this, we developed a structured SVM which predicts
  all components jointly, using the union of the features of each
  component. We use approximate inference, first enumerating possible
  trigger lists, and then equation trees, and find the best scoring
  structure. However, this method did not outperform the pipeline
  method. The worse performance of joint learning is due to: (1)
  search space being too large for the joint model to do well given
  our dataset size of 385, and (2) our independent classifiers being
  good enough, thus supporting better joint inference. This tradeoff
  is strongly supported in the literature \cite{PRYZ05,SuttonMc07}.

  Another option is to enforce constraints between trigger list
  predictions, such as, variable triggers should not overlap with the
  quantity triggers.  However, we noticed that often noun phrases
  returned by the Stanford parser were noisy, and would include
  neighboring numbers within the extracted noun phrases. This
  prevented us from enforcing such constraints.  

\section{Experimental Results}
  We now describe the data set, and the annotation procedure used. We
  then evaluate the system's performance on predicting trigger list,
  equation tree, and the complete equation parse.

\subsection{Dataset}\label{sec:dataset} 

  We created a new dataset consisting of $385$ sentences extracted
  from algebra word problems and financial news headlines. For algebra
  word problems, we used the MIT dataset \cite{KushmanZeBa14}, and two
  high school mathematics textbooks, Elementary Algebra (College of
  Redwoods) and Beginning and Intermediate Algebra (Tyler
  Wallace). Financial news headlines were extracted from The Latest
  News feed of MarketWatch, over the month of February, 2015.
  {\em All} sentences with information
  describing a mathematical relation among at most two (possibly
  coreferent) variables, were chosen. Next, we pruned sentences which
  require multiple uses of a number to create the equation. This only
  removed a few time related sentences like {\em ``In 10 years, John
    will be twice as old as his son.''}. We empirically found that
  around 97\%
  of sentences describing a relation fall under the scope of our dataset.




  The annotators were shown each sentence paired with the normalized
  equation representing the relation in the sentence.  For each
  variable in the equation, the annotators were asked to mark spans of
  text which best describe what the variable represents. The
  annotation guidelines are provided in the appendix.  We wanted to
  consider only noun phrase constituents for variable
  grounding. Therefore, for each annotated span, we extracted the noun
  phrase with maximum overlap with the span, and used it to represent
  the variables. Finally, a tuple with each variable being mapped to
  one of the noun phrases representing it, forms a valid output
  grounding (variable trigger list). We computed inter-annotator
  agreement on the final annotations where only noun phrases represent
  variables. The agreement (kappa) was 0.668, indicating good
  agreement. The average number of mention annotations per sentence
  was 1.74.

\subsection{Equation Parsing Modules}
  
  In this section, we evaluate the performance of the individual
  modules of the equation parsing process. We report Accuracy - the
  fraction of correct predictions. Table \ref{tab:modules} shows the
  $5$-fold cross validation accuracy of the various modules. In each
  case, we also report accuracy by removing each feature group, one at
  a time. In addition, for equation tree prediction, we also show the
  effect of lexicon, projectivity, conforming to syntactic parse
  constraints, and using lexicon as features instead of hard
  constraints.  For all our experiments, we use the Stanford Parser
  \cite{SBMN13}, the Illinois POS tagger \cite{RothZe98} and the
  Illinois-SL structured prediction package \cite{ChangUCSR15}.
  \begin{table}
    \centering
    \footnotesize
    \begin{tabular}{|l|c|}
    \hline
    Quantity Trigger List Prediction & Accuracy\\\hline\hline
    All features & \textbf{95.3} \\\hline
    No Neighborhood features & 42.5 \\\hline
    No Quantity features & 93.2 \\\hline
    \multicolumn{2}{l}{} \\\hline
    Variable Trigger List Prediction & Accuracy\\\hline\hline
    All features & \textbf{75.5} \\\hline
    No Variable features & 58.6 \\\hline
    No Neighborhood features & 70.3 \\\hline
    \multicolumn{2}{l}{} \\\hline
    Equation Tree Prediction & Accuracy \\\hline\hline
    All features & \textbf{78.9} \\\hline
    No Neighborhood features & 64.3 \\\hline
    No Connecting Text features & 70.2 \\\hline
    No Number features & 77.6 \\\hline
    No Lexicon & 72.7 \\\hline 
    No Projectivity & 72.8 \\\hline
    Conform with Syntactic Parse & 70.2 \\ \hline
    Lexicon as Features & 74.5 \\ \hline
    \end{tabular}
    \caption{\footnotesize  Performance of system components}
    \label{tab:modules}
  \end{table}
  \setlength{\tabcolsep}{6pt}
      \begin{table}[!ht]
        \centering \footnotesize
        \begin{tabular}{|l|C{2cm}|C{2cm}|}
          \hline 
           Source & Equation Accuracy & Equation + Grounding Accuracy\\\hline
          \hline 
           Our System & \textbf{71.3} & \textbf{61.2} \\\hline
           Joint Learning & 60.9 & 50.0 \\\hline 
           SPF & 3.1 & N/A \\
          \hline
        \end{tabular}
        \caption{\footnotesize Performance on equation parsing}
        \label{tab:fullsystem}
      \end{table}

\subsection{Equation Parsing Results}

  In this section, we evaluate the performance of our system on the
  overall equation parsing task. We report Equation Accuracy - the
  fraction of sentences for which the system got the equation correct,
  and Equation+Grounding Accuracy - the fraction of sentences for
  which the system got both the equation and the grounding of
  variables correct. Table \ref{tab:fullsystem} shows the overall
  performance of our system, on a $5$-fold cross validation. We
  compare against Joint Learning - a system which jointly learns to
  predict all relevant components of an equation parse (Section
  \ref{sec:alternatives}). We also compare with SPF \cite{ArtziZe13},
  a publicly available semantic parser, which can learn from
  sentence-logical form pairs. We train SPF with sentence-equation
  pairs and a seed lexicon for mathematical terms (similar to ours),
  and report equation accuracy.  Our structured predictors pipeline
  approach is shown to be superior to both Joint Learning and SPF.

  SPF gets only a few sentences correct. We attribute this to the
  inability of SPF to handle overlapping mentions (like in Example 4),
  as well as its approach of parsing the whole sentence to the final
  output form. The developers of SPF also confirmed~\footnote{Private
  communication} that it is not suitable for equation parsing and that these results are expected.  
  %
  Since equation parsing is a more involved process, a slight adaptation of
  SPF does not seem possible, necessitating a more involved
  process , of the type we propose. Our approach, in contrast to SPF, can handle overlapping
  mentions, selects triggers from text, and parses the trigger
  list to form equations.

\subsection{Error Analysis}
  
  For variable trigger list prediction, around $25\%$ of the errors
  were due to the predictor choosing a span which is contained within
  the correct span, e.g., when the target noun phrase is ``The
  cost of a child's ticket'', our predictor chose only ``child's
  ticket''. Although this choice might be sufficient for downstream
  tasks, we consider it to be incorrect in our current
  evaluation. Another $25\%$ of the errors were due to selection of
  entities which do not participate in the relation.  For example, in
  ``A rancher raises 5 times as many cows as horses.'', our predictor
  chose ``A rancher'' and ``cows'' as variables, whereas the relation
  exists between ``cows'' and ``horses''. For the prediction of the
  equation tree, we found that $35\%$ of the errors were due to rare
  math concepts expressed in text. For example, ``7 dollars short
  of the price'' represents $7$ dollars should be subtracted from the
  price. These errors can be handled by carefully augmenting the
  lexicon. Another $15\%$ of the errors were due to lack of world
  knowledge, requiring understanding of time, speed, and distance.

\section{Conclusion}
  This paper investigates methods that identify and understand
  mathematical relations expressed in text. We introduce the equation
  parsing task, which involves generating an equation from a sentence
  and identifying what the variables represent. We define the notion
  of projectivity, and construct a high precision lexicon, and use these to reduce the equation search space.  Our experimental
  results are quite satisfying and raise a few interesting
  issues. In particular, it suggests that predicting equation parses
  using a pipeline of structured predictors performs better than
  jointly trained alternatives. As discussed, it also points out the
  limitation of the current NLP tools in supporting these tasks.  Our
  current formulation has one key limitation; we only deal with
  expressions that are described within a sentence. Our future work
  will focus on lifting this restriction, in order to allow relations
  expressed across multiple sentences and multiple relations expressed
  in the same sentence. Code and dataset are available
  at \url{http://cogcomp.cs.illinois.edu/page/publication\_view/800}.

\section*{Acknowledgements}
This work is funded by DARPA under agreement number FA8750-13-2-0008, and a grant from the Allen
Institute for Artificial Intelligence (allenai.org).

\appendix
\section*{Appendix}
\section{Features}

\subsection{Quantity Trigger List Prediction}

The feature function $\phi(x,y)$ used for the classification generates
the following features :

    \begin{enumerate}
      \item \textbf{Neighborhood features} : For each quantity $q$ in
        the input sentence, we add unigrams and bigrams generated from
        a window around $q$, part of speech tags of neighborhood
        tokens of $q$. We conjoin these features with $\Relevance(q)$.

      \item \textbf{Quantity Features} : For each quantity $q$, we add
        unigrams and bigrams of the phrase representing the quantity.
        Also, we add a feature indicating whether the number is
        associated with number one or two, and whether it is the only
        number present in the sentence. These features are also
        conjoined with $\Relevance(q)$.
    \end{enumerate}

\subsection{Variable Trigger List Prediction}

The features $\phi(x,y)$ used for variable
    trigger prediction are as follows:
    \begin{enumerate}
      \item \textbf{Variable features} : Unigrams and bigrams
        generated from the noun phrase representing variables, part of
        speech tags of tokens in noun phrase representing
        variables.

      \item \textbf{Neighborhood Features} : Unigrams and POS tags
         from neighborhood of variables.

    \end{enumerate}
    All the above features are conjoined with two labels, one denoting
    whether $y$ has two variables or one, and the second denoting
    whether $y$ has two variables represented by the same noun phrase.

\subsection{Equation Tree Prediction}

For features $\phi(x,y)$, we extract for each non-leaf node $n$ of
    the equation tree $y$, the following:
    \begin{enumerate}
      \item \textbf{Neighborhood Features} : Unigrams, bigrams and POS
        tags from neighborhood of $\SpanStart(lc(n))$,
        $\SpanStart(rc(n))$, $\SpanEnd(lc(n))$ and $\SpanEnd(rc(n))$,
        conjoined with $\Op(n)$ and $\Order(n)$.

      \item \textbf{Connecting Text Features} : Unigrams, bigrams and
        POS tags between $\min(\SpanEnd(lc(n)),\SpanEnd(rc(n)))$ and
        $\max(\SpanStart(lc(n)),\SpanStart(rc(n)))$, conjoined with
        $\Op(n)$ and $\Order(n)$.

      \item \textbf{Number Features} : In case we are combining two leaf
        nodes representing quantity triggers, we add a feature signifying
        whether one number is larger than the other.
    \end{enumerate}

\section{Annotation Guidelines}

  The annotators were shown each sentence paired with the normalized
  equation representing the relation in the sentence.  For each
  variable in the equation, the annotators were asked to mark spans of
  text which best describe what the variable represents. They were
  asked to annotate associated entities if exact variable description
  was not present. For instance, in example 3 (Section 1), the
  relation holds between the speed of bird and the speed of
  wind. However, {\em``speed''} is not explicitly mentioned in the
  sentence.  In such cases, the annotators were asked to annotate the
  associated entities {\em``the wind''} and {\em``a bird''} as
  representing variables.

  The guidelines also directed annotators to choose the longest
  possible mention, in case they feel the mention boundary is
  ambiguous. As a result, in the sentence, {\em``City Rentals rent an
  intermediate-size car for 18.95 dollars plus 0.21 per mile.''}, the
  phrase {\em``City Rentals rent an intermediate-size car''} was
  annotated as representing variable. We allow multiple mentions to be
  annotated for the same variable. In example 2 (Section 1), both
  {\em``a number''} and {\em``the same number''} were annotated as
  representing the same variable.

\section{Lexicon}

  We construct a high precision list of rules, to parse sentences
  describing mathematical concepts, for example, ``difference of'',
  ``greater than'', etc. For each non-leaf node $n$ of a projective
  equation tree, we define the following terms :

  \begin{enumerate}
    \item $\MidSpan(n)$ : The string from $\min(\SpanEnd(lc(n)), 
    \SpanEnd(rc(n)))$ to $\max(\SpanStart(lc(n)), \SpanStart(rc(n)))$.

    \item $\LeftSpan(n)$ : The string ending at
    $\min(\SpanStart(lc(n)), \SpanStart(rc(n)))$ and starting from the
    nearest trigger position on the left.

    \item $\RightSpan(n)$ : The string starting at
    $\max(\SpanEnd(lc(n)), \SpanEnd(rc(n)))$ and ending at the nearest
    trigger position on the right.
  
    \item $\LeftToken(n)$ : Defined only for leaves, indicates the
    span of text for the trigger of $n$.

  \end{enumerate}
  
   The rules in our lexicon are described using the above terms. They
   are as follows, ordered from low precedence to high precedence.

   \begin{enumerate}

     \item If $\LeftSpan(n)$ contains ``sum of'' and $\MidSpan(n)$
     contains ``and'' or is the empty string, $\Op(n)$ should be $+$.

     \item If $\MidSpan(n)$ contains one of ``added to'', ``plus'',
     ``more than'' ``taller than'', ``greater than'', ``larger than'',
     ``faster than'', ``longer than'', ``increased'', $\Op(n)$ should
     be $+$.

     \item If $\MidSpan(n)$ contains one of ``more than'' ``taller
     than'', ``greater than'', ``larger than'', ``faster than'',
     ``longer than'', and $\RightSpan(n)$ contains ``by'', $\Op(n)$
     should be $-$, and $\Order(n)$ should be $lr$.

     \item If $\LeftSpan(n)$ contains ``difference of'' and
     $\MidSpan(n)$ contains ``and'' or is the empty string, $\Op(n)$
     should be $-$ and $\Order(n)$ should be $lr$.

     \item If $\LeftSpan(n)$ contains one of ``exceeds'', ``minus'',
     ``decreased'', $\Op(n)$ should be $-$, and $\Order(n)$ should be
     $lr$.

     \item If $\MidSpan(n)$ contains one of ``subtracted'' ``shorter
     than'', ``less than'', ``slower than'', ``smaller than'',
     $\Op(n)$ should be $-$, and $\Order(n)$ should be $rl$.

     \item If $\MidSpan(n)$ contains ``multiplied by'', $\Op(n)$ is
     $\times$.

     \item If $\LeftSpan(n)$ contains ``product of'' and $\MidSpan(n)$
     contains ``and'', $\Op(n)$ should be $\times$.

     \item If $\LeftSpan(n)$ contains ``ratio of'', $\Op(n)$ should be
     $\div$, and $\Order(n)$ should be $lr$.

     \item If $\LeftToken(n)$ contains one of ``thrice'', ``triple'',
     ``twice'', ``double'', ``half'', or if $\MidSpan(n)$ contains
     ``times'', $\Op(n)$ is $\times$.

     \item If $\LeftToken(n)$ contains one of ``thrice'', ``triple'',
     ``twice'', ``double'', ``half'', or if $\MidSpan(n)$ contains
     ``times'', and $\MidSpan(n)$ contains ``as'', and $\RightSpan(n)$
     contains ``as'', operation at $\Op(n)$ is $\div$, and $\Order(n)$
     is $rl$.

   \end{enumerate}

\bibliographystyle{emnlp2016}
\bibliography{ccg,cited,equation}

\end{document}